%% file: neurips_2026.tex
\documentclass{article}
\usepackage[preprint]{neurips_2026}

\usepackage[utf8]{inputenc} %
\usepackage[T1]{fontenc}    %
\usepackage{hyperref}       %
\usepackage{url}            %
\usepackage{booktabs}       %
\usepackage{amsfonts}       %
\usepackage{nicefrac}       %
\usepackage{microtype}      %
\usepackage{xcolor}         %

\usepackage[textsize=tiny]{todonotes}
\usepackage{multirow}
\usepackage{graphicx}
\usepackage{caption}
\usepackage{subcaption}
\usepackage{longtable}
\usepackage{hhline}

\usepackage{tikz}
\usetikzlibrary{arrows.meta, positioning, fit, backgrounds, calc}

\usepackage{float}
\usepackage{algorithm}
\usepackage{algorithmic}
\usepackage{enumitem}

\usepackage{tabularx}
\usepackage{url}     



\usepackage{amsmath}
\usepackage{amssymb}
\usepackage{wrapfig} 
\usepackage{rotating}
\usepackage{mathtools}
\usepackage{subcaption}
\usepackage{amsthm}
\usepackage{tcolorbox}
\usepackage[normalem]{ulem}

\usepackage{natbib}

\DeclareMathOperator*{\argmax}{arg\,max}
\DeclareMathOperator*{\argmin}{arg\,min}

\usepackage{dsfont}

\title{Advancing Weighted Partial MaxSAT Solving via Differentiable GPU Guidance}

\author{Mingju Liu$^{1}$ \quad Curie Kim$^{1}$ \quad Steve Dai$^{2}$ \quad Cunxi Yu$^{1,3}$ \\ $^{1}$University of Maryland, College Park, MD, USA \\ $^{2}$NVIDIA Research, Santa Clara, CA, USA \\ $^{3}$NVIDIA Research, College Park, MD, USA}

\makeatletter
\renewcommand{\paragraph}{\@startsection{paragraph}{4}{\z@}%
  {0.5ex \@plus 0.4ex \@minus .2ex}%
  {-1em}%
  {\normalfont\normalsize\bfseries}}
\makeatother

\begin{document}

\maketitle

\begin{abstract}
We present a GPU-CPU hybrid framework for Weighted Partial MaxSAT (WPMAX-SAT) that complements state-of-the-art CPU solvers with differentiable GPU optimization. The GPU stage directly optimizes instance-specific categorical decision parameters without a training dataset or a pretrained predictive model. The central challenge in extending differentiable optimization to WPMAX-SAT is the hard/soft clause asymmetry: hard clauses represent inviolable feasibility constraints while soft clauses carry weights. We introduce a \textbf{hierarchical differentiable loss} that prioritizes hard clause satisfaction before optimizing the weighted soft objective, ensuring that GPU gradient descent leans towards the constraint structure of WPMAX-SAT. Our framework transforms the GPU-optimized variable logits to high-confidence assignments and transfers them to CPU solvers via two complementary modes: \textbf{partial guidance}, which injects the most confident perturbed partial variable assignments as soft clauses into $N$ parallel solver instances, and \textbf{phase initialization}, which guides the CDCL solver's internal branching polarity with the full GPU assignment as a warm start to preserve the problem's originality. We evaluate on all 229 instances of the MSE 2024 (most recent due to cancellation in 2025) anytime weighted benchmark with the top 2 winners as baselines; our hybrid framework improves the average solution score by up to \textbf{4.9\%} over \textit{tt-open-wbo-inc-Glucose} and \textbf{4.8\%} over \textit{SPB-MaxSAT-c-FPS} at a 300-second time limit, with the potential to obtain better solutions surpassing the best known solution recorded in the competition evaluation.
\end{abstract}

\input{1-Intro}
\input{2-Preliminary}
\input{3-Methodology}

\input{4-Experiments}
\input{5-Conclusion}

\bibliography{bib/references}
\bibliographystyle{plainnat}

\appendix
\vspace{-0.8em}

\section{Selected Instance Problem Size and Score}\label{ap:score}

\begin{table}[ht]
\centering
\caption{20 selected instances based on average score improvement across all methods and runtime limits. 
\#V, \#H, \#S denote the number of variables, hard clauses, and soft clauses in the original WCNF. Shown scores are recorded at 300\,s.}
\label{tab:top20}
\setlength{\tabcolsep}{3.5pt}
\resizebox{\linewidth}{!}{%
\begin{tabular}{l|rrr|ccc|ccc}
\hline
\textbf{Instance} & \textbf{\#V} & \textbf{\#H} & \textbf{\#S}
  & \textbf{\textit{TT}} & \textbf{\textit{TT}-PG} & \textbf{\textit{TT}-PI} &
  \textbf{\textit{SPB}} & \textbf{\textit{SPB}-PG} & \textbf{\textit{SPB}-PI} \\
\hline
DT-TicTacToe-L127-D6-AB5  & 278K  & 2527K &  765  & 0.610 & 1.405 & 1.092 & 0.881 & \textbf{2.063} & 1.256 \\
DT-Heart-CL-L127-D6-AB5   & 869K  & 7032K &  236  & 0.537 & 1.175 & 1.288 & 0.573 & \textbf{1.446} & 0.734 \\
DT-TicTacToe-L127-D6-AB3  & 278K  & 2527K &  765  & 0.773 & 1.351 & 0.882 & 0.843 & \textbf{1.648} & 1.261 \\
Syn-R2-16-S9               & 433K  & 1297K &  935  & 0.532 & \textbf{1.075} & 0.853 & 0.593 & 0.913 & 1.000 \\
SwitchAct-OR1200-66        & 532K  & 1940K &   66  & \textbf{1.000} & \textbf{1.000} & \textbf{1.000} & 0.385 & \textbf{1.000} & \textbf{1.000} \\
DT-Heart-CL-L15-D3-AB6    &  19K  &  125K &  236  & 0.717 & \textbf{1.129} & 1.101 & 0.723 & 0.902 & 0.967 \\
Syn-R2-10-S11              & 581K  & 1741K & 1034  & 0.862 & 0.854 & 1.002 & 0.699 & \textbf{1.384} & 1.075 \\
DT-PrimTumor-L63-D5-AB4   &  82K  &  660K &  268  & 0.735 & 0.962 & 0.776 & 0.772 & 0.912 & \textbf{1.179} \\
DT-PrimTumor-L127-D6-AB4  & 312K  & 2556K &  268  & 0.456 & 0.703 & 0.578 & 0.382 & 0.542 & \textbf{0.716} \\
SwitchAct-OR1200-200       & 532K  & 1941K &  200  & 0.973 & 1.200 & 0.973 & 0.956 & 1.059 & \textbf{1.301} \\
Syn-R2-17-S10              & 1138K & 3409K & 1261  & 0.830 & 0.965 & 0.885 & 0.678 & 0.945 & \textbf{1.076} \\
SwitchAct-OR1200-100       & 532K  & 1940K &  100  & 0.840 & 1.000 & \textbf{1.105} & 0.840 & 0.840 & 1.000 \\
SwitchAct-OR1200-70        & 532K  & 1940K &   70  & \textbf{1.000} & 0.778 & \textbf{1.000} & 0.636 & \textbf{1.000} & \textbf{1.000} \\
Railway-wg2                & 112K  & 2877K & 2236  & 0.790 & \textbf{1.000} & \textbf{1.000} & 0.975 & \textbf{1.000} & \textbf{1.000} \\
Timetabling-comp07         & 217K  & 1805K & 2293  & \textbf{1.000} & 0.538 & \textbf{1.000} & 0.538 & \textbf{1.000} & \textbf{1.000} \\
Syn-R1-14-S6               & 249K  &  747K &  786  & 0.951 & \textbf{1.266} & 0.995 & 0.943 & 0.995 & 0.943 \\
Syn-R2-11-S10              & 651K  & 1951K & 1187  & 0.549 & 0.762 & 0.549 & 0.767 & \textbf{0.817} & 0.767 \\
QC-Grover-7                &  25K  &  298K & 4641  & \textbf{1.000} & \textbf{1.000} & \textbf{1.000} & \textbf{1.000} & \textbf{1.000} & \textbf{1.000} \\
Timetabling-comp10         & 150K  & 1212K & 1987  & \textbf{1.000} & \textbf{1.000} & \textbf{1.000} & \textbf{1.000} & \textbf{1.000} & \textbf{1.000} \\
AbsRef-polysite-bloat      & 11.7M & 26.9M & 5986  & 0.784 & 0.426 & 0.909 & 0.833 & 0.444 & \textbf{1.081} \\
\hline
\end{tabular}}
\end{table}

\section{Multi-thread/seed Score Distribution at 60s: Partial Guidance/Phase Initialization}\label{ap:dist}
\begin{figure}[H]
    \centering
    \includegraphics[width=\linewidth]{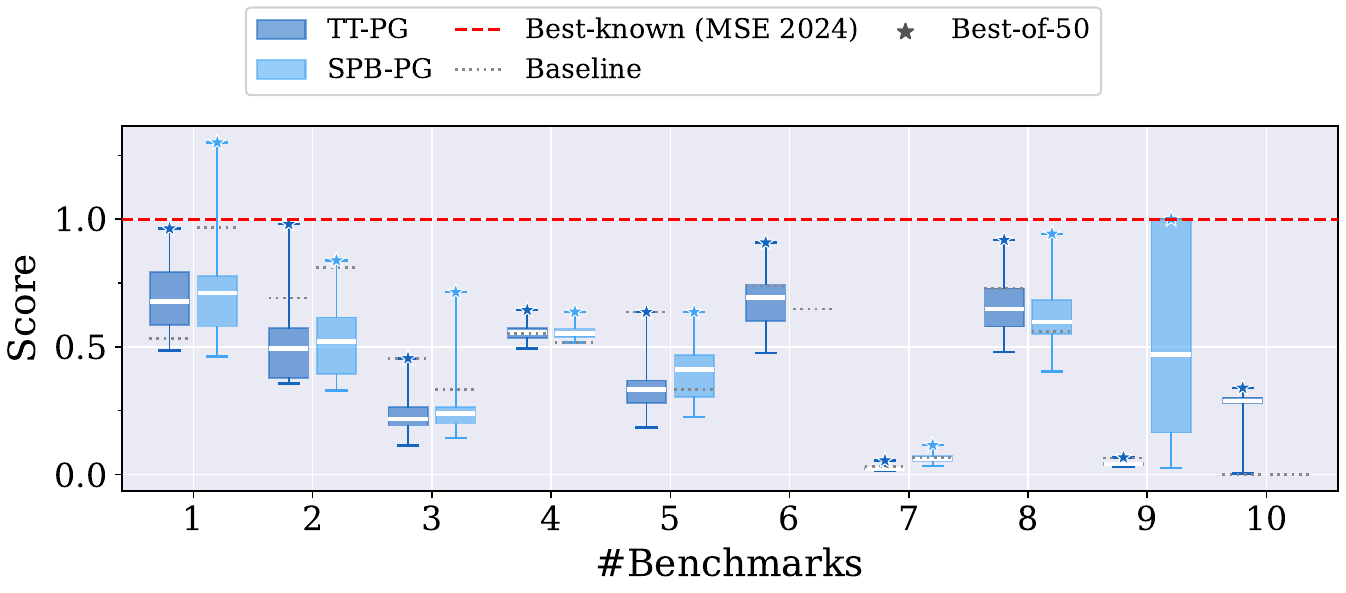}
    \caption{Per-instance score distribution of 50 Gumbel-perturbed guidance thread scores at 60\,s for 10 selected instances. Each box spans the IQR; the white line marks the median; whiskers extend to the min/max of all 50 scores; the star ($\star$) marks the best-of-50 result; the dotted line marks the single-thread original baseline.}
    \label{fig:pg_thread_dist_60}
\end{figure}

\begin{figure}[H]
    \centering
    \includegraphics[width=\linewidth]{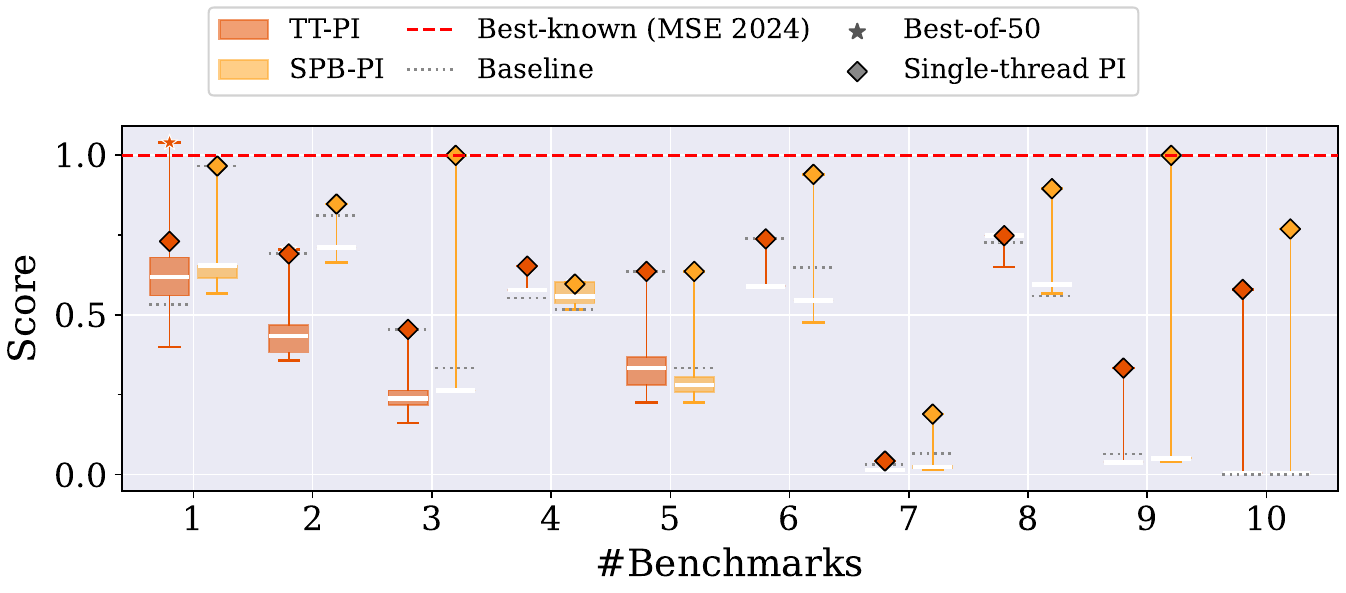}
    \caption{Per-instance distribution of 50 seed scores at 60\,s for the same 10 instances as Figure~\ref{fig:pg_thread_dist_60}. Each box spans the IQR; the white line marks the median; whiskers extend to the min/max of all 50 scores; the star ($\star$) marks the best-of-50 result; the diamond ($\diamond$) marks the single-thread PI score; the dotted line marks the single-thread original baseline.}
    \label{fig:pi_seed_dist_60}
\end{figure}

Figure. \ref{fig:pg_thread_dist_60} and Figure. \ref{fig:pi_seed_dist_60} show the score distribution at 60s to complement the full results in Figure. \ref{fig:pg_thread_dist_300} and Figure. \ref{fig:pi_seed_dist_300}, respectively. 

\section{Selected Instance Full Name}
Table~\ref{tab:abbre} lists the full benchmark instance names corresponding to the abbreviations used in Table~\ref{tab:top20}.

\small
\setlength{\tabcolsep}{5pt}
\renewcommand{\arraystretch}{1.2}
\begin{longtable}{l|p{0.62\linewidth}}
\caption{Full names for the 20 selected instances in Table~\ref{tab:top20}.}
\label{tab:abbre}\\
\hline
\textbf{Abbreviation} & \textbf{Full Instance Name} \\
\hline
\endfirsthead
\hline
\textbf{Abbreviation} & \textbf{Full Instance Name} \\
\hline
\endhead
\hline
\endfoot
DT-TicTacToe-L127-D6-AB5  & \path{decision-tree-tic-tac-toe-un-formula_0.8_2021_atleast_127_max-6_reduced_incomplete_adaboost_5} \\
DT-Heart-CL-L127-D6-AB5   & \path{decision-tree-heart-cleveland-un-formula_0.8_2021_atleast_127_max-6_reduced_incomplete_adaboost_5} \\
DT-TicTacToe-L127-D6-AB3  & \path{decision-tree-tic-tac-toe-un-formula_0.8_2021_atleast_127_max-6_reduced_incomplete_adaboost_3} \\
Syn-R2-16-S9              & \path{synplicate_new-synplicate_synthetic_dag_run2_16_size_9_ms_26_50_0_erm} \\
SwitchAct-OR1200-66       & \path{switchingactivitymaximization-SwitchingActivityMaximization_OpenRISC1200_66} \\
DT-Heart-CL-L15-D3-AB6    & \path{decision-tree-heart-cleveland-un-formula_0.8_2021_atleast_15_max-3_reduced_incomplete_adaboost_6} \\
Syn-R2-10-S11             & \path{synplicate_new-synplicate_synthetic_dag_run2_10_size_11_ms_26_50_0_erm} \\
DT-PrimTumor-L63-D5-AB4   & \path{decision-tree-primary-tumor-un-formula_0.8_2021_atleast_63_max-5_reduced_incomplete_adaboost_4} \\
DT-PrimTumor-L127-D6-AB4  & \path{decision-tree-primary-tumor-un-formula_0.8_2021_atleast_127_max-6_reduced_incomplete_adaboost_4} \\
SwitchAct-OR1200-200      & \path{switchingactivitymaximization-SwitchingActivityMaximization_OpenRISC1200_200} \\
Syn-R2-17-S10             & \path{synplicate_new-synplicate_synthetic_dag_run2_17_size_10_ms_26_50_0_erm} \\
SwitchAct-OR1200-100      & \path{switchingactivitymaximization-SwitchingActivityMaximization_OpenRISC1200_100} \\
SwitchAct-OR1200-70       & \path{switchingactivitymaximization-SwitchingActivityMaximization_OpenRISC1200_70} \\
Railway-wg2               & \path{railway-transport_wt-wg2} \\
Timetabling-comp07        & \path{timetabling_wt-comp07} \\
Syn-R1-14-S6              & \path{synplicate_new-synplicate_synthetic_dag_run1_14_size_6_ms_0_25_0_erm} \\
Syn-R2-11-S10             & \path{synplicate_new-synplicate_synthetic_dag_run2_11_size_10_ms_26_50_0_erm} \\
QC-Grover-7               & \path{quantum-circuit-grover-noancilla_4_52_ibmq-casablanca_7} \\
Timetabling-comp10        & \path{timetabling_wt-comp10} \\
AbsRef-polysite-bloat     & \path{abstraction-refinement_wt-polysite-bloat} \\
\end{longtable}

\newpage

\end{document}

%% file: 1-Intro.tex
\section{Introduction}\label{introduction}

The Boolean satisfiability problem (SAT)~\cite{hansen1990algorithms} determines whether a logical formulation has a satisfying assignment; its optimization complement, Maximum Satisfiability (MaxSAT), determines the assignment that satisfies as many clauses as possible. In its practically important \textit{Weighted Partial MaxSAT} (WPMAX-SAT) variant, a subset of clauses is designated as \textit{hard}---they must be satisfied in any valid solution---while the remaining \textit{soft} clauses carry positive integer weights. The objective is to find a feasible assignment that satisfies all hard clauses while minimizing the total weight of violated soft clauses. This hard/soft structure directly captures the trade-off between inviolable constraints and weighted constraints that arises across a wide spectrum of real applications. In electronic design automation (EDA), MaxSAT underlies post-routing timing optimization, technology mapping, and automatic test pattern generation (ATPG)~\cite{drechsler2008acceleration}.
In planning and scheduling, it encodes resource constraints as hard clauses while treating objective penalties as soft clauses~\cite{zhang2012maxsat}. In formal verification, MaxSAT captures optimal fault localization and minimum-cost repair problems~\cite{orvalho2025counterexample}. The breadth of these domains reflects a fundamental inspiration: most real-world combinatorial optimization problems are not pure feasibility questions---they are optimization problems with hard constraints, and WPMAX-SAT is their fundamental basis.

The MaxSAT Evaluation (MSE)~\cite{mse2024} drives critical progress in MaxSAT solving, complementing the SAT Competition~\cite{SATCompetitionWeb} in SAT. The top-two winners in the MSE 2024 anytime weighted track, \textit{tt-open-wbo-inc-Glucose}~\cite{nadel2019anytime,nadel2020polarity,nadel2024tt} and \textit{SPB-MaxSAT-c-FPS}~\cite{cai2023nuwls,zheng2024rethinking,jiang2025enhancing}, both rely on a conflict-driven clause learning (CDCL) engine at their core. These solvers are fundamentally sequential---each branching decision and conflict-learning step is taken in series---and the massive parallelism offered by modern GPUs/TPUs is largely left unexploited.

A prior work~\cite{kim2025galoissat} showed that differentiable GPU optimization can meaningfully accelerate CDCL-based SAT solving. At its core, the observation is: SAT solving can be seen as a differentiable optimization problem by encoding Boolean logic through pseudo-Boolean algebraic representations and applying a differentiable loss that \textit{maximizes the number of satisfied clauses}---which is MaxSAT-style. The variable assignments produced by this GPU optimization are then used as guidance to narrow the CDCL search space. This raises a direct question: if the GPU optimization in that prior work is already built on a MaxSAT-style loss function, can the same framework be extended to WPMAX-SAT?
The challenge is that the existing uniform clause-counting loss treats all clauses equally: it assigns the same cost to violating a hard constraint as to violating a soft clause, making it incompatible with the hard/soft asymmetry of WPMAX-SAT without re-implementation.

This paper introduces a GPU-guided framework that extends the hybrid GPU-CPU paradigm of~\cite{dai2025turbosat, kim2025galoissat} to the WPMAX-SAT setting by addressing this gap precisely. The critical technical contribution is a \textbf{hierarchical loss function} that encodes the hard/soft constraint asymmetry directly into the GPU optimization objective: hard clause violations incur a penalty strictly greater than the maximum achievable soft reward, ensuring that the optimizer always prioritizes feasibility before optimizing the weighted soft objective. From the GPU differentiable optimization, our framework extracts high-confidence variable assignments and transfers them to CPU MaxSAT solvers via two complementary guidance modes. In \textbf{partial guidance}, confidence-ranked variable assignments are injected as soft clauses into the WCNF problem, utilizing $N$ parallel solver threads toward high-quality regions of the search space without restricting their completeness. In \textbf{phase initialization}, the full GPU solved assignments are fed into the CDCL solver's internal branching polarity before any search begins, providing a global warm start while preserving the completeness and originality of the MaxSAT instances.

The differentiable stage is training-free and dataset-free: its logits are optimization variables for the current WCNF instance, and gradients are computed from that instance's clauses and weights. No model is fitted across a collection of instances. Repeated gradient updates are part of solving the instance, rather than a separate model-training phase.

The key design principle of our framework is that it operates as a \textit{complement} to any existing MaxSAT solvers. For CDCL-based solvers, the backbone retains its complete and deterministic algorithmic power while our framework provides a high-quality starting point derived from GPU-parallel optimization over the problem structure.
Evaluated on the MSE 2024 anytime weighted track benchmark using both top-two competition solvers \textit{tt-open-wbo-inc-Glucose} and \textit{SPB-MaxSAT-c-FPS}, our framework consistently improves solution quality across both at 60\,s and 300\,s runtime limits on the full 229-instance benchmark.

The main contributions of this work are:
\begin{itemize}[nosep]
    \item A \textbf{hierarchical differentiable loss} for WPMAX-SAT that strictly prioritizes hard clause feasibility before optimizing the weighted soft objective, enabling GPU gradient-based optimization to respect the hard/soft constraint asymmetry (Section~\ref{gpu_main}).
    \item \textbf{Partial guidance via soft clause injection}: top-$\rho$ most confident variable assignments from $N$ independently Gumbel-perturbed samples are injected as soft clauses into $N$ copies of the original WCNF, running in parallel and returning the best solution. Applied to the 229 instances of MSE 2024 anytime weighted benchmark at 300s runtime limit, partial guidance improves the average score by up to \textbf{4.9\%} over the two baseline solvers. (Section~\ref{partial_guidance}).
    \item \textbf{Phase initialization with optional multi-seed parallelism}: the full GPU assignment seeds the CDCL solver's initial branching phase while leaving the original WCNF unaugmented. Optionally, we utilize $N$ threads to diversify the search through independent random seeds for both baseline solvers. Phase initialization improves the average solution score by up to \textbf{2.6\%} over the baseline while preserving the problem's originality (Section~\ref{phase_init}).
\end{itemize}

%% file: 2-Preliminary.tex
\section{Background}\label{background}

\paragraph{Weighted Partial Maximum Satisfiability}

Maximum Satisfiability (MaxSAT) is an optimization extension of the Boolean Satisfiability problem (SAT). Whereas SAT asks whether all clauses have any satisfying assignment, MaxSAT asks for the assignment that satisfies as many clauses as possible. Both problems are NP-complete or NP-hard~\cite{karp2009reducibility}. In the \textit{Weighted Partial MaxSAT} (WPMAX-SAT) formulation, the clause set is partitioned into \textit{hard} clauses $\mathcal{H}$ that must be satisfied by any valid solution, and \textit{soft} clauses $\mathcal{S} = \{(C_t, w_t)\}$ each carrying a positive integer weight $w_t$. The objective is to find a Boolean assignment satisfying all hard clauses while minimizing the total weight of violated soft clauses. This hard/soft asymmetry makes WPMAX-SAT the natural formulation for optimization problems with inviolable constraints and weighted constraints, and it arises across a broad range of domains: EDA for timing optimization, technology mapping, and ATPG; planning and scheduling for resource constraint encoding; and formal verification for minimum-cost repair and fault localization. 

\paragraph{Modern MaxSAT Solvers}

SOTA MaxSAT solving is dominated by two paradigms. \textbf{CDCL-based solvers} build on conflict-driven clause learning originally developed for SAT~\cite{marques2002grasp}, interleaving SAT oracle calls with bound propagation to enumerate improving solutions. \textit{Open-WBO}~\cite{martins2014openwbo} established a modular CDCL-based MaxSAT framework; \textit{tt-open-wbo-inc-Glucose} extended it with anytime upper-bound strategies, employing CDCL to enforce hard clause feasibility and integrating progressive bound propagation to continuously improve solution quality within a runtime limit; it ranked second in the MSE 2024 anytime weighted track at both runtime limits. \textbf{Hybrid SLS solvers} combine stochastic local search with an embedded CDCL engine. \textit{NuWLS-c}~\cite{cai2023nuwls} introduced clause-weight update strategies that significantly improve solution quality under time constraints. \textit{SPB-MaxSAT-c-FPS} builds on \textit{NuWLS-c} as its local search core and also embeds Glucose~\cite{glucose2009} for hard clause completeness; the local search component flips variable assignments guided by weighted clause heuristics while Glucose handles the CDCL proof obligations. It ranked first in the same MSE 2024 track at both time limits. Both solvers are fundamentally sequential in their core decision loops, leaving the parallelism of modern GPU hardware largely unexploited.

\paragraph{GPU-Guided SAT Solving}

Prior works~\cite{dai2025turbosat,kim2025galoissat} demonstrated that differentiable GPU optimization can accelerate CDCL-based SAT solving. It encodes Boolean clause satisfaction through a pseudo-Boolean algebraic representation and applies a Gumbel-Softmax straight-through estimator to optimize a uniform clause-counting objective over batched variable assignments in parallel on a GPU. The resulting assignments guide a CDCL solver by narrowing its search space. However, this prior work treats all clauses identically due to the SAT solving nature: violating a hard constraint incurs the same cost as violating the soft clause, making it incompatible with the hard/soft asymmetry of WPMAX-SAT. Our framework directly addresses this gap with a hierarchical loss that prioritizes hard feasibility before optimizing the weighted soft objective.

%% file: 3-Methodology.tex
\section{Methodology}\label{methodology}

\begin{figure}[ht]
\centering
\resizebox{\textwidth}{!}{\input{Figures/figure_arch_1}}  
\caption{
Overview of our framework's architecture and solving flow.
(a) \textbf{GPU Differentiable Optimization}. The differentiable GPU optimization maintains a logit vector per variable and evaluates hard and soft clause satisfaction in a unified forward pass. A hierarchical loss prioritizes hard clause satisfaction besides optimizing the weighted soft objective.
(b) \textbf{Partial Guidance Mode}. Confidence-ranked variable assignments from $N$ independently Gumbel-perturbed samples are injected as soft guidance clauses into the original WCNF, enabling parallel guided search across $N$ CPU solver threads.
(c) \textbf{Phase Initialization Mode}. The full GPU assignment seeds the initial internal branching phase of the CDCL-based solver. Optionally, a variant utilizes $N$ threads to run the CPU solver using the same initialization with distinct random seeds to diversify search paths.
}
\vspace{-2mm}
\label{fig:framework}
\end{figure}

Our framework extends the hybrid GPU-CPU paradigm of~\cite{kim2025galoissat} to the MaxSAT solving domain. Whereas the prior work addresses the decision problem of SAT, our framework targets WPMAX-SAT, a significantly more complex and fundamental combinatorial optimization problem that requires balancing hard feasibility constraints against a weighted soft-clause objective. The framework uses GPU parallelism to run differentiable optimization over all variable assignments simultaneously, driving a hierarchical objective that leans towards enforcing hard constraints while maximizing weighted soft clause satisfaction, then transfers the assignments as guidance to deterministic CPU based MaxSAT solvers.

Formally, a WPMAX-SAT instance is given as a Weighted CNF (WCNF) formula $\varphi = (\mathcal{H}, \mathcal{S})$ over $n$ Boolean variables, where $\mathcal{H}$ denotes the set of hard clauses that must be satisfied by any valid solution, and $\mathcal{S} = \{(C_t, w_t)\}$ is the set of soft clauses, each carrying a positive integer weight $w_t$. The objective is to find an assignment $\mathbf{x} \in \{0,1\}^n$ that satisfies all hard clauses and minimizes the total weight of violated soft clauses:
\begin{equation}\label{eq.maxsat_obj}
    \min_{\mathbf{x} \in \{0,1\}^n} \quad \sum_{t \in \mathcal{S}} w_t \cdot \mathds{1}[C_t(\mathbf{x}) = 0], \quad \text{subject to } C_t(\mathbf{x}) = 1 \;\;\forall\, t \in \mathcal{H}.
\end{equation}

As illustrated in Fig.~\ref{fig:framework}(a), the GPU module vectorizes all $n$ variables as logit vectors and optimizes them through a hierarchical differentiable loss function that prioritizes hard clause feasibility while maximizing the weighted soft clause coverage. After optimization, the optimized logits yield a distribution over variable assignments from which candidate solutions are extracted and transferred to the CPU solving stage via one of two guidance modes. In \textbf{partial guidance} (Fig.~\ref{fig:framework}b), per-variable probability from the Gumbel-Softmax distribution inspires the selection of the most confident assignments, which are injected as soft clauses into $N$ copies of the original WCNF, and these augmented problems are solved with $N$ parallel solver threads. In \textbf{phase initialization} (Fig.~\ref{fig:framework}c), the full GPU assignment seeds the CDCL solver's internal branching phase before any search begins, providing a warm start without modifying the problem. In both modes, the CPU solvers perform complete MaxSAT searches and return the solution with the best objective value.

\subsection{Differentiable MaxSAT Formulation}\label{gpu_main}

\paragraph{Variable Representation and Clause Evaluation}

Following~\cite{kim2025galoissat}, each Boolean variable $x_i$ is parameterized by a logit vector $\boldsymbol{\theta}_i = [\theta_{i,0}, \theta_{i,1}] \in \mathbb{R}^2$, representing unnormalized scores for the two Boolean states. In practice, these logits are maintained in batched form as $\Theta \in \mathbb{R}^{B \times n \times 2}$, where $B$ is the batch size and each of the $B$ samples is an independent candidate assignment, all randomly initialized to explore alternative assignments in parallel. These logits are instance-specific optimization variables rather than parameters of a predictive model. They are updated directly using the current instance's objective, without a training dataset.

The satisfaction of each clause $C_t$ is expressed through the pseudo-Boolean algebraic encoding of disjunction originally introduced in~\cite{kim2025galoissat}: $C_t(\hat{\mathbf{x}}) = 1 - \prod_{i \in C_t}(1 - s_{t,i}), \; C_t \in \{0, 1\},$
where $s_{t,i} \in \{0, 1\}$ is the Boolean evaluation of the $i$-th literal in clause $C_t$: specifically, $s_{t,i} = \hat{x}_j$ if the literal is positive and $s_{t,i} = 1 - \hat{x}_j$ if negative, with $j$ denoting the underlying variable index. The product equals zero if and only if at least one literal is satisfied, so $C_t = 1$ whenever the clause is satisfied and $C_t = 0$ otherwise.

\paragraph{Gradient-based Optimization with Gumbel-Softmax}

Purely soft relaxation of the discrete assignments, as shown in the logic vector, distorts clause satisfaction values: a clause with two equally likely variables each assigned probability $0.5$ would yield a relaxed satisfaction of $1 - 0.5^2 = 0.75$, misrepresenting the actual probability of satisfaction. To resolve this, we adopt the straight-through (ST) Gumbel-Softmax formulation, which applies the Gumbel-Softmax reparameterization~\cite{gumbel1954statistical,jang2017categoricalreparameterizationgumbelsoftmax,maddison2017concretedistributioncontinuousrelaxation} to enable differentiable sampling from a categorical distribution. Independent Gumbel noise $g_{b,i,j}$ is added to each logit, followed by a temperature-controlled softmax:
\begin{equation}\label{eq.gumbel}
\mathbf{y}_{b,i,j} = \frac{\exp\!\bigl((\theta_{b,i,j} + g_{b,i,j})/\tau\bigr)}{\displaystyle\sum_{t \in \{0,1\}} \exp\!\bigl((\theta_{b,i,t} + g_{b,i,t})/\tau\bigr)},
\quad g_{b,i,j} \stackrel{\text{i.i.d.}}{\sim} \mathrm{Gumbel}(0,1),
\end{equation}
where $\tau > 0$ is the temperature parameter. In the forward pass, we apply the ST estimator to obtain a discrete assignment: $\hat{x}_{b,i} = \argmax_j \, \mathbf{y}_{b,i,j},\; \hat{x}_{b,i} \in \{0,1\},$
while the backward pass substitutes its gradient with that of the soft relaxation, i.e., $\partial\mathcal{L}/\partial\hat{x}_{b,i} \approx \partial\mathcal{L}/\partial\mathbf{y}_{b,i}$. This straight-through approximation allows gradients to flow through the non-differentiable $\argmax$, so that the framework optimizes on discrete clause evaluations without sacrificing differentiability.

\paragraph{Hierarchical Loss for Weighted Partial MaxSAT}

A fundamental challenge in extending differentiable optimization to WPMAX-SAT is the hard/soft clause asymmetry: hard clauses represent Boolean feasibility constraints, while soft clauses define a weighted optimization objective. A uniform loss that aggregates all clauses as in~\cite{kim2025galoissat} would allow the optimizer to trade hard constraint violations for soft clause gains, yielding infeasible solutions that may harm the guidance quality. To enforce the hard clause priority, we introduce a differentiable \textbf{hierarchical loss} that strictly penalizes hard clause violations while simultaneously rewarding weighted soft clause satisfaction:
\begin{equation}\label{eq.loss}
    \mathcal{L}^{(b)} = \underbrace{\lambda \cdot \Big(|\mathcal{H}| - \sum_{t \in \mathcal{H}} C_t\!\bigl(\hat{\mathbf{x}}^{(b)}\bigr)\Big)}_{\text{hard clause penalty}} \;-\; \underbrace{\sum_{t \in \mathcal{S}} w_t \cdot C_t\!\bigl(\hat{\mathbf{x}}^{(b)}\bigr)}_{\text{soft clause reward}},
\end{equation}
where $|\mathcal{H}|$ is the total number of hard clauses and $\lambda$ is a penalty coefficient.
In standard WCNF encoding, the top weight $W_{\text{top}}$ is defined as $W_{\text{top}} = 1 + \sum_{t \in \mathcal{S}} w_t$, so that setting $\lambda = \alpha \cdot W_{\text{top}}$ with $\alpha \geq 1$ guarantees: $\lambda \geq W_{\text{top}} = 1 + \sum_{t \in \mathcal{S}} w_t \;>\; \sum_{t \in \mathcal{S}} w_t,$
making the cost of a single hard clause violation strictly greater than the maximum possible soft reward, regardless of the problem size. The batched forward pass objective is the mean of $\mathcal{L}^{(b)}$ over all $B$ batch samples, minimized jointly through gradient descent.

After the differentiable optimization, the batch sample $b^*$ achieving the minimum hierarchical loss is selected for solution extraction: $b^* = \argmin_{b \in \{1,\ldots,B\}} \mathcal{L}^{(b)},$
and its associated logits $\theta_{\text{sel}} = \Theta_{b^*}$ are retained as the initialization for the subsequent CPU solving stage. This selection identifies the sample that best balances hard feasibility and soft optimality under the hierarchical objective.

\subsection{Guided Solving Flow}

\subsubsection{Partial Guidance via Soft Clause Injection}\label{partial_guidance}

Starting from the best-batch logits $\theta_{\text{sel}}$, we generate $N$ candidate variable assignments using the same Gumbel-Softmax process as in optimization, but with independent noise draws. For each candidate $k \in \{0, \ldots, N-1\}$, we compute the Gumbel-perturbed logits and extract an assignment: $\tilde{\boldsymbol{\theta}}_i^{(k)} = \boldsymbol{\theta}_{\text{sel},i} + \mathbf{g}_i^{(k)}, \; \hat{x}_i^{(k)} = \argmax_j \; \tilde{\theta}_{i,j}^{(k)},$
where $\mathbf{g}_i^{(0)} = \mathbf{0}$ for the deterministic reference sample and $g_{i,j}^{(k)} \stackrel{\text{i.i.d.}}{\sim} \mathrm{Gumbel}(0,1)$ for $k = 1, \ldots, N-1$. The confidence of the assignment for variable $i$ in sample $k$ is the probability margin of the Gumbel-Softmax output: $c_i^{(k)} = \bigl[\mathrm{softmax}(\tilde{\boldsymbol{\theta}}_i^{(k)})\bigr]_{\hat{x}_i^{(k)}},$
which measures how decisively the perturbed logits favor the assigned state. The $N-1$ stochastic samples diversify the assignment pool; together with the deterministic reference, they provide $N$ assignments that share the same high-quality logits but explore different variable decisions.

For each candidate $k$, we rank all $n$ variables by $c_i^{(k)}$ and select the top $\rho$ fraction as guidance targets: $S^{(k)} \subseteq \{1,\ldots,n\}$ with $|S^{(k)}| = \lceil \rho n \rceil$. Only the most confident variable decisions are selected, while uncertain variables are left unconstrained. Each selected variable contributes a unit soft clause aligned with its predicted value: 
\begin{equation}\label{eq.soft_inject}
    \ell_i^{(k)} = \begin{cases} x_i, & \hat{x}_i^{(k)} = 1, \\ \lnot x_i, & \hat{x}_i^{(k)} = 0, \end{cases}
\end{equation}
yielding one augmented formulation per thread: $\varphi_{\text{aug}}^{(k)} = \varphi \;\wedge\; \bigwedge_{i \in S^{(k)}} \bigl(w_g,\; \ell_i^{(k)}\bigr),$
where $w_g$ is the guidance weight, set as a fixed fraction of the minimum soft clause weight in $\varphi$. Since all injected clauses are soft, the solver may violate any of them if doing so produces a lower objective or is needed to satisfy a hard constraint. The $N$ augmented instances are distributed to $N$ CPU solver threads running in parallel, and the solution with the minimum objective across all threads is recorded as the final optimized objective.

\subsubsection{Phase Initialization}\label{phase_init}

As a complementary guidance strategy, our framework uses the GPU assignment to initialize the branching phase of the CDCL-based MaxSAT solvers. The full deterministic assignment is extracted from the optimized logits: $\hat{\mathbf{x}}_{\text{full}} = \argmax_j \; [\mathrm{softmax}(\theta_{\text{sel}})]_j \;\in\; \{0,1\}^n$
and written into the solver's internal phase array before the first CDCL call, so that all initial branching decisions start from the GPU-suggested assignment. Unlike partial guidance, which augments the WCNF with additional soft clauses and operates on a modified problem, phase initialization leaves the original problem formulation $\varphi$ completely unchanged. The solver receives the unmodified WCNF and is only given a warm-start assignment as its initial branching polarity; clause learning will naturally revise any poor initial decision, and the solver retains full freedom to reach any part of the search space. This makes phase initialization lean towards the original problem, while still providing high-quality guidance for warm-start from the GPU optimization. Notably, for hybrid SLS-based solvers such as \textit{SPB-MaxSAT-c-FPS} that embed a CDCL engine as their core search component, the same phase initialization mechanism can be applied uniformly.

\paragraph{Multi-seed Parallel Phase Initialization Variant}

To further explore, we introduce a \textbf{multi-seed} variant in which $N$ solver processes share the same phase initialization $\hat{\mathbf{x}}_{\text{full}}$ but are each assigned a distinct random seed $s_k = s_0 + k$ for $k = 0, \ldots, N-1$. The random seed controls solver's internal stochastic decisions, such as random restarts and tie-breaking in branching heuristics, so the $N$ processes collectively explore a diverse set of search paths sharing the same initialization. The solution achieving the minimum objective across all $N$ threads is reported as the optimized result.

%% file: Figures/figure_arch_1.tex
\begin{tikzpicture}[
  >=Stealth,
  every node/.style={font=\normalsize},
  io/.style={
    rectangle, draw=black!70, fill=white,
    text width=3.6cm, align=center, minimum height=0.8cm},
  proc/.style={
    rectangle, rounded corners=3pt, draw=black!70, fill=gray!12,
    text width=3.6cm, align=center, minimum height=0.8cm},
  math/.style={
    rectangle, rounded corners=3pt, draw=blue!55!black, fill=blue!8,
    text width=3.6cm, align=center, minimum height=0.8cm},
  branchbox/.style={
    rectangle, rounded corners=3pt, draw=black!70, fill=gray!12,
    text width=2.5cm, align=center, minimum height=0.8cm, font=\normalsize},
  lossbox/.style={
    rectangle, rounded corners=3pt, draw=orange!70!black, fill=orange!12,
    text width=3.6cm, align=center, minimum height=0.8cm},
  outbox/.style={
    rectangle, rounded corners=3pt, draw=green!55!black, fill=green!12,
    text width=3.6cm, align=center, minimum height=0.8cm},
  arr/.style={->, thick, black!80},
  backarr/.style={->, thick, dashed, black!45},
  xarr/.style={->, thick, dotted, black!80},
  caption_style/.style={font=\normalsize\bfseries, align=center, text width=6.5cm}
]

\begin{scope}[node distance=0.45cm]
\node[io]   (wcnf)   {WCNF $\varphi$};
\node[proc, below=of wcnf]   (part)   {Hard/Soft Partition\\$(\mathcal{H},\;\mathcal{S})$};
\node[math, below=of part]   (theta)  {Logit Matrix $\Theta$\\$\in\mathbb{R}^{B\times n\times 2}$};
\node[math, below=of theta]  (gumbel) {Gumbel-Softmax\\$Y\!\in\![0,1]^{B\times n\times 2}$};
\node[math, below=of gumbel] (st)     {Straight-Through\\$\hat{X}\!\in\!\{0,1\}^{B\times n}$};

\node[branchbox] (hpen) at ($(st.south)+(-1.85cm,-1.3cm)$) {%
  \textbf{Hard Penalty}\\
  $|\mathcal{H}|-\!\sum_{t\in\mathcal{H}}\!C_t(\hat{x})$};
\node[branchbox] (srew) at ($(st.south)+(+1.85cm,-1.3cm)$) {%
  \textbf{Soft Reward}\\
  $\sum_{t\in\mathcal{S}} w_t C_t(\hat{x})$};

\node[lossbox, below=2.6cm of st] (lossfn) {%
  Hierarch.\ Loss\\
  $\mathcal{L}\!=\!\lambda P_h(\hat{X}) - R_s(\hat{X})$};
\node[outbox, below=of lossfn] (tsel_a) {%
  Best-Batch Select.\\
  $\theta_\mathrm{sel}\!\in\!\mathbb{R}^{n}$};
\end{scope}

\draw[arr] (wcnf)   -- (part);
\draw[arr] (part)   -- (theta);
\draw[arr] (theta)  -- (gumbel);
\draw[arr] (gumbel) -- (st)
  node[midway, right=2pt, font=\normalsize] {fwd};
\draw[backarr] (gumbel.west) to[bend left=55]
  node[midway, left=2pt, font=\normalsize] {$\nabla$\,bwd} (theta.west);
\draw[arr] (st.south) -- (hpen.north);
\draw[arr] (st.south) -- (srew.north);
\draw[arr] (hpen.south) |- (lossfn.west);
\draw[arr] (srew.south) |- (lossfn.east);
\draw[arr] (lossfn) -- (tsel_a);

\node[caption_style, below=0.35cm of tsel_a] {(a) GPU Differentiable Optimization};

\begin{scope}[node distance=0.85cm]
\node[outbox] (tsel_b) at (8.5cm, 0) {$\theta_\mathrm{sel}$};
\node[math,  below=of tsel_b]  (gpert)   {%
  Gumbel Perturbation\\
  $\tilde{\theta}^{(k)}\!=\!\theta_\mathrm{sel}+g^{(k)}$};
\node[proc,  below=of gpert]   (conf)    {%
  Confidence-Ranked\\
  Variable Select.\ $S^{(k)}$};
\node[proc,  below=of conf]    (cands)   {%
  $N$ Assignments\\
  $\hat{x}^{(0)},\ldots,\hat{x}^{(N-1)}$};
\node[proc,  below=of cands]   (augwcnf) {%
  Augmented WCNF\\
  $\varphi\cup\{(w_g,\,\ell_i^{(k)})\}_{i\in S^{(k)}}$};
\node[proc,  below=of augwcnf] (cpub)    {%
  $N$ Parallel CPU\\Solver Threads};
\end{scope}

\node[outbox]   (bestb)  at (tsel_b |- tsel_a) {%
  Best Solution\\(min.\ objective)};

\draw[arr] (tsel_b)  -- (gpert);
\draw[arr] (gpert)   -- (conf);
\draw[arr] (conf)    -- (cands);
\draw[arr] (cands)   -- (augwcnf);
\draw[arr] (augwcnf) -- (cpub);
\draw[arr] (cpub)    -- (bestb);

\node[caption_style, below=0.35cm of bestb] {(b) Partial Guidance Mode};

\begin{scope}[node distance=1.3cm]
\node[outbox] (tsel_c) at (15.5cm, 0) {$\theta_\mathrm{sel}$};
\node[math,  below=of tsel_c]     (fullx)  {%
  Full Assignment\\
  $\hat{x}_\mathrm{full}\!=\!\mathrm{argmax}(\theta_\mathrm{sel})$};
\node[proc,  below=of fullx]      (phase)  {%
  CDCL Phase/\\Polarity Seeding};
\node[proc,  below=of phase]      (seeds)  {%
  Multi-Seed Parallel\\
  $s_k\!=\!s_0\!+\!k,\;k\!=\!0,\ldots,N{-}1$};
\node[proc,  below=of seeds]      (cpuc)   {%
  $N$ Parallel CPU\\Solver Threads};
\end{scope}

\node[outbox]      (bestc) at (tsel_c |- tsel_a) {%
  Best Solution\\(min.\ objective)};

\draw[arr] (tsel_c) -- (fullx);
\draw[arr] (fullx)  -- (phase);
\draw[arr] (phase)  -- (seeds);
\draw[arr] (seeds)  -- (cpuc);
\draw[arr] (cpuc)   -- (bestc);

\node[fit=(seeds)(cpuc), draw=red!60, dashed, inner sep=6pt, rounded corners=2pt] (optbox) {};
\node[font=\normalsize\bfseries, red!80!black, fill=white, inner sep=2pt]
  at ($(seeds.south)!0.5!(cpuc.north)$) {Optional};

\node[caption_style, below=0.35cm of bestc] {(c) Phase Initialization Mode};

\coordinate (fork) at ($(tsel_a.east) + (2.4cm, 0)$);
\coordinate (forkB) at (fork |- tsel_b.west);
\coordinate (forkC) at (fork |- {$(tsel_b.north)+(0, 0.7cm)$});

\draw[thick, dotted, black!80] (tsel_a.east) -- (fork) -- (forkC);
\draw[xarr] (forkB) -- (tsel_b.west);
\draw[xarr] (forkC) -| (tsel_c.north);

\begin{scope}[on background layer]
  \node[fit=(wcnf)(tsel_a)(hpen)(srew),
        inner xsep=12pt, inner ysep=10pt,
        draw=gray!45, dashed, rounded corners=6pt,
        fill=gray!4] {};
  \node[fit=(tsel_b)(bestb),
        inner xsep=12pt, inner ysep=10pt,
        draw=blue!40, dashed, rounded corners=6pt,
        fill=blue!3] {};
  \node[fit=(tsel_c)(bestc),
        inner xsep=12pt, inner ysep=10pt,
        draw=teal!50!black!40, dashed, rounded corners=6pt,
        fill=teal!3] {};
\end{scope}

\end{tikzpicture}

%% file: 4-Experiments.tex
\section{Experiments}\label{experiments}

\paragraph{Benchmark}
We evaluate our framework on the MSE 2024 anytime weighted track benchmark, which contains 229 WPMAX-SAT instances -- the most up-to-date evaluation set. To fit our GPU differentiable formulation, we apply the Tseitin transformation to convert each instance to 3MaxSAT.

\paragraph{Solvers and Baselines}
We use the top-two winners from the MSE 2024 anytime weighted track as backbone solvers: \textit{tt-open-wbo-inc-Glucose} (\textit{TT}) and \textit{SPB-MaxSAT-c-FPS} (\textit{SPB}). For each solver, the baseline is the unmodified, single-threaded solver run with only CPU. We evaluate both guidance modes: \textbf{partial guidance} (\textit{TT}-PG and \textit{SPB}-PG) and \textbf{phase initialization} (\textit{TT}-PI and \textit{SPB}-PI), at runtime limits of \textbf{60\,s} and \textbf{300\,s}, which align with the MSE 2024 anytime track settings. We conduct all experiments on an Intel\textregistered{} Xeon\textregistered{} Gold 6418H CPU and an NVIDIA RTX\texttrademark{} A6000 GPU.

\paragraph{Metric}
We adopt the MSE 2024 anytime scoring formula: $\mathrm{score}(v) = \frac{\mathrm{Best_{sol}} + 1}{\mathrm{obj}(v) + 1},$
where $\mathrm{Best_{sol}}$ is the best objective value achieved by any solver in the official MSE 2024 competition, and $\mathrm{obj}(v)$ is the achieved objective value. Note that a score above 1.0 is possible: it indicates that our solver finds a solution better than any result recorded in the official MSE 2024 competition.
We report the average score across all 229 instances, with each method's score floored at the corresponding single-thread baseline (i.e., a single-thread baseline always runs in parallel as a fallback).

\paragraph{Hyperparameters}
The GPU differentiable optimization runs for 10 epochs with gradient accumulation, using a target batch size of 250 and a fixed Gumbel-Softmax temperature of $\tau=1$. For partial guidance, we use $N = 50$ parallel solver threads; the guidance fraction is $\rho = 0.0002$ (top 0.02\% most confident variables), and the partial guidance weight $w_g$ is set to a fixed 10\% of the minimum soft clause weight in the WCNF.

\subsection{Main Results}\label{exp:main}

\begin{table}[ht]
\centering
\vspace{-0.3cm}
\caption{Average solution score across all 229 MSE 2024 anytime weighted instances at 60\,s and 300\,s runtime limits. \textbf{Bold} marks the best-performing configuration. $\Delta$ reports absolute improvement over the respective baseline. Values are reported as \textit{TT}/\textit{SPB}.}
\label{tab:main_results}
\resizebox{\linewidth}{!}{%
\begin{tabular}{l|cc|cc}
\hline
\multirow{2}{*}{\textbf{Solver Configuration}} &
\multicolumn{2}{c|}{\textbf{60\,s}} &
\multicolumn{2}{c}{\textbf{300\,s}} \\
& \textbf{Score} & $\Delta$ & \textbf{Score} & $\Delta$ \\
\hline
\textit{TT}/\textit{SPB} (single-thread)             & 0.866/0.876 & ---/---         & 0.925/0.926 & ---/---         \\
\textit{TT}/\textit{SPB}-PI (single-thread)          & 0.883/0.900 & +0.017/+0.024   & 0.944/0.950 & +0.019/+0.024   \\
\textbf{\textit{TT}/\textit{SPB}-PG} (multi-thread $N$=50) & \textbf{0.903/0.910} & \textbf{+0.037/+0.034} & \textbf{0.970/0.970} & \textbf{+0.045/+0.044} \\
\hline
\end{tabular}}
\vspace{-0.2cm}
\end{table}

Table~\ref{tab:main_results} summarizes the average solution scores for all six methods on the full 229-instance benchmark. PG-powered solvers deliver the largest gains: \textit{TT}-PG and \textit{SPB}-PG both reach an average score of \textbf{0.970} at 300\,s, compared to 0.925 and 0.926 for the respective baselines, achieving a $+$4.9\% and $+$4.8\% scoring gain, respectively. PI-powered solvers provide more modest yet consistent gains, reaching 0.944 (\textit{TT}-PI) and 0.950 (\textit{SPB}-PI) at 300\,s, a $+$2.1\% and $+$2.6\% improvement. At 60\,s, both PG- and PI-powered solvers achieve over 2.0\% scoring gain over their respective baselines, confirming that the GPU guidance is effective even at the shorter runtime limit.

\begin{figure}[ht]
\centering
\includegraphics[width=0.49\linewidth]{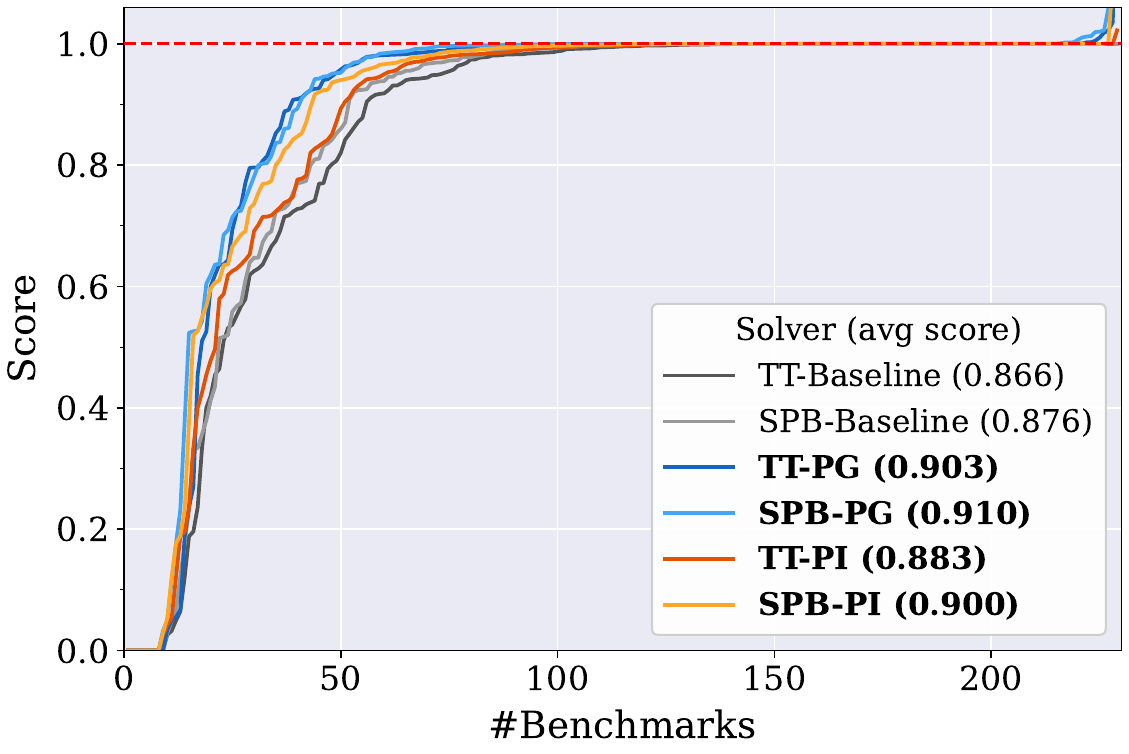}%
\hfill%
\includegraphics[width=0.49\linewidth]{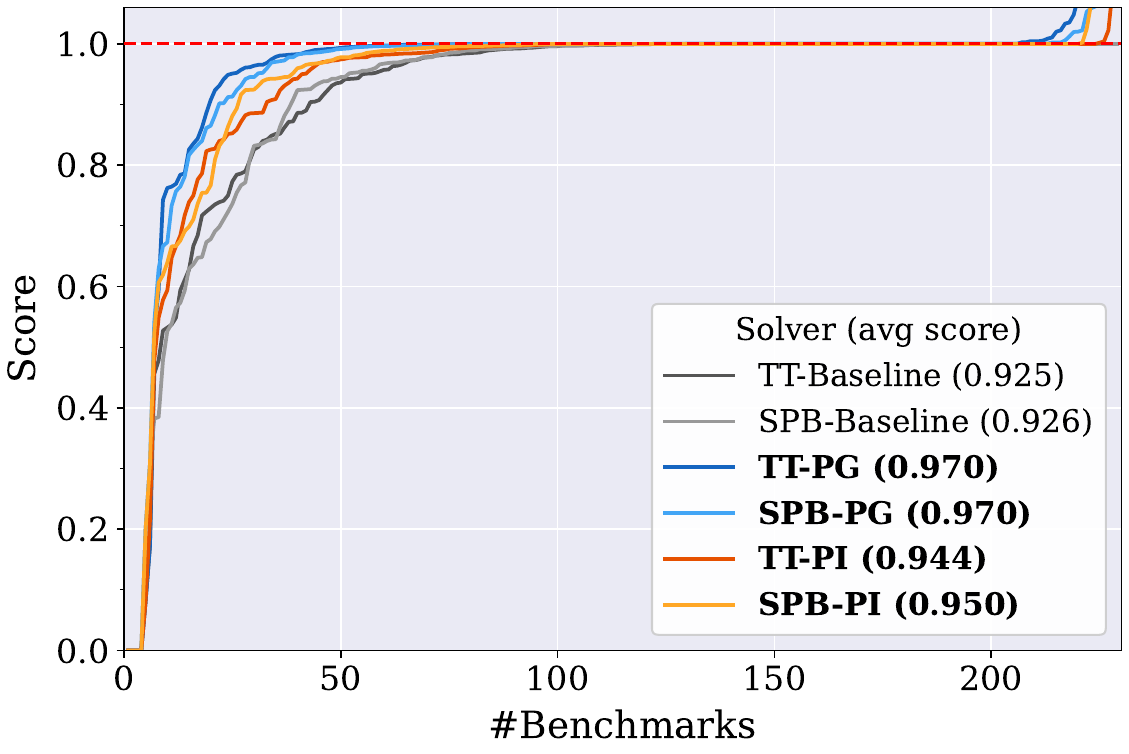}
\caption{Per-instance sorted score on all 229 MSE 2024 anytime weighted instances at 60\,s (left) and 300\,s (right). The red dashed line marks a score of 1.0 (best-known MSE 2024 result). Both guidance modes consistently outperform their respective baselines across all instances, with partial guidance providing the largest improvement.}
\label{fig:main_anytime}
\end{figure}

Figure~\ref{fig:main_anytime} shows the per-instance score trend across all 229 instances. Across both solvers and both runtime limits, all four guided methods progressively dominate their corresponding baselines, confirming that GPU guidance consistently improves solution quality. On over 5\% of the instances at both runtime limits, the score exceeds 1.0, indicating that our guided solvers find solutions better than any known best solution in the MSE 2024 competition, setting a new standard in the competition.

\subsection{Ablation Studies}\label{exp:ablation}

To study how GPU guidance translates across independent solver runs and different solving paths, we select the 20 instances where our framework achieves the largest average score improvement over the baseline across all guidance modes and solvers, based on the results in Section~\ref{exp:main}. We list these instances with their problem sizes and per-method scores at 300\,s in Table~\ref{tab:top20} in Appendix~\ref{ap:score}. 

For a closer look, we pick 10 of the 20 instances
that together showcase diverse multi-thread and multi-seed score distributions across both guidance modes and both runtime limits. For each instance, we plot the score distribution as a box spanning the interquartile range (IQR), with the entire range extending to the full min/max score across all 50 threads or seeds.

\subsubsection{Multi-thread Score Distribution: Partial Guidance}\label{exp:ablation_pg}

\begin{figure}[ht]
    \centering
    \vspace{-0.2cm}
    \includegraphics[width=0.9\linewidth]{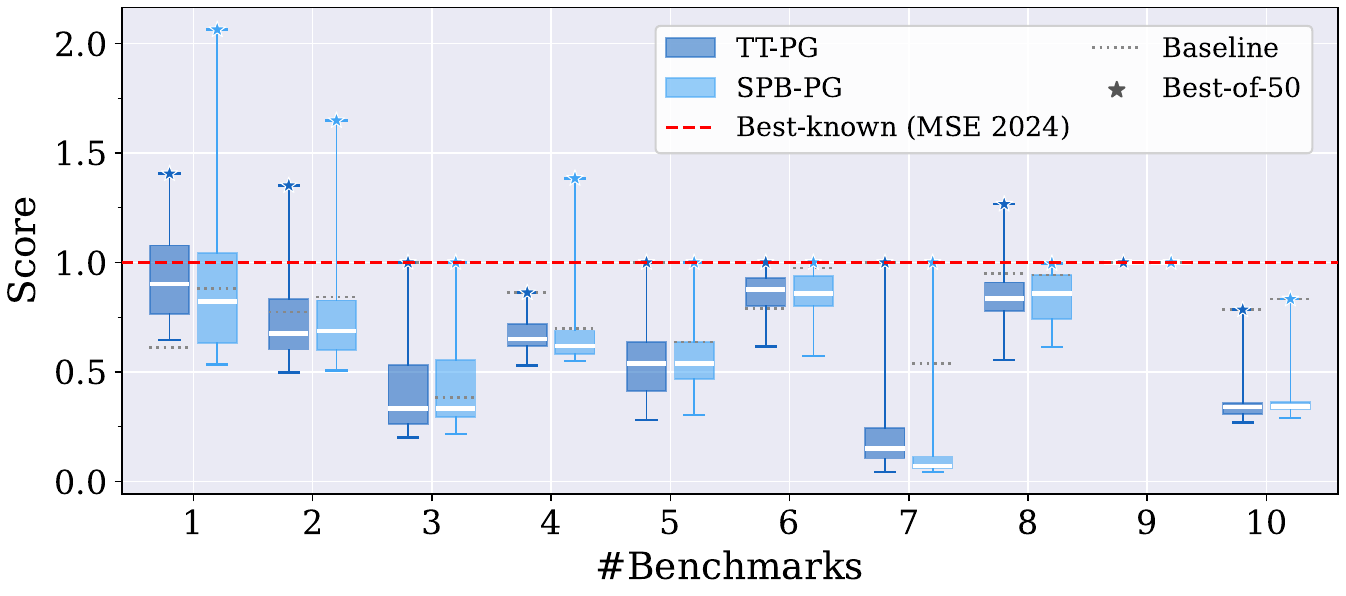}
    \caption{Per-instance score distribution of 50 Gumbel-perturbed guidance thread scores at 300\,s for 10 selected instances. Each box spans the IQR; the white line marks the median; whiskers extend to the min/max of all 50 scores; the star ($\star$) marks the best-of-50 result; the dotted line marks the single-thread original baseline. (60\,s results in Appendix~\ref{ap:dist}.)}
    \label{fig:pg_thread_dist_300}
    \vspace{-0.2cm}
\end{figure}

With $N = 50$ parallel threads each receiving an independently Gumbel-perturbed guidance from the same optimized logits $\theta_\mathrm{sel}$, Figure~\ref{fig:pg_thread_dist_300} shows the score distribution per instance at 300s runtime limits. Each instance exhibits a wide spread across its 50 threads: on average, \textit{TT}-PG achieves a per-thread score of 0.696 with a best-of-50 of 1.405, while \textit{SPB}-PG averages 0.671 with a best-of-50 of 2.063. The large gap between the per-thread average and the best-of-50 motivates the parallel design: individual Gumbel perturbations vary widely in quality, and multi-threading with best-thread selection reliably captures a high-quality guidance sample. On instance~\#1, \textit{SPB}-PG reaches a best-of-50 score of 2.063 while its per-thread median remains well below the best known solution. On instance~\#8, \textit{TT}-PG improves from a baseline of 0.951 to 1.266, whereas the \textit{SPB}-PG cannot reach a better solution than the best known solution at the same runtime limit---showing that the benefit of Gumbel guidance is solver-dependent and not uniform across different methods.

\subsubsection{Multi-seed Score Distribution: Phase Initialization}\label{exp:ablation_pi}

\begin{figure}[ht]
    \centering
    \vspace{-0.1cm}
    \includegraphics[width=0.9\linewidth]{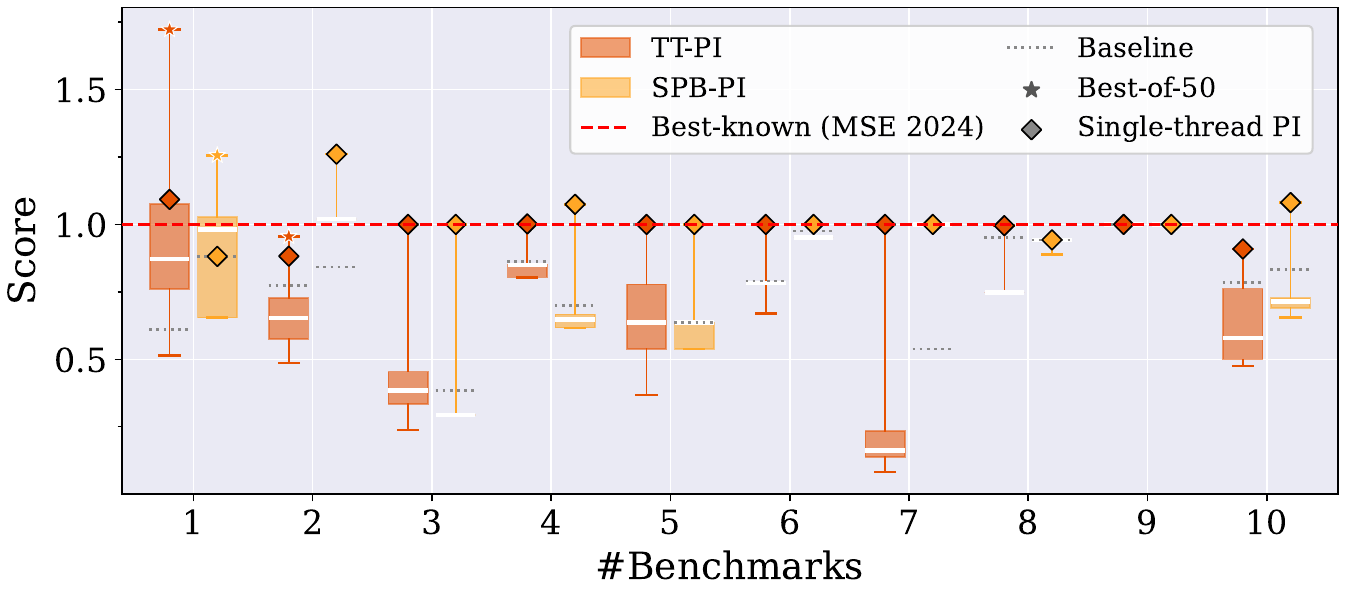}
    \caption{Per-instance distribution of 50 seed scores at 300\,s for the same 10 instances as Figure~\ref{fig:pg_thread_dist_300}. Each box spans the IQR; the white line marks the median; whiskers extend to the min/max of all 50 scores; the star ($\star$) marks the best-of-50 result; the diamond ($\diamond$) marks the single-thread PI score; the dotted line marks the single-thread original baseline. (60\,s results in Appendix~\ref{ap:dist}.)}
    \label{fig:pi_seed_dist_300}
    \vspace{-0.2cm}
\end{figure}

For phase initialization, we run $N = 50$ solver processes, each sharing the same GPU assignment $\hat{\mathbf{x}}_\mathrm{full}$: 49 processes each use a distinct random seed from 0 to 48, and one process uses the default solver seed. Figure~\ref{fig:pi_seed_dist_300} shows the per-seed score distribution per instance at 300s runtime limits. Compared to partial guidance, the distributions are more compact and stable: \textit{TT}-PI averages 0.715 per seed with best-of-50 of 1.722, while \textit{SPB}-PI averages 0.770 with best-of-50 of 1.256. The tighter range reflects that all 50 seeds share the same GPU full warm-start, so the search space is consistently biased toward the GPU-suggested assignment. Different seeds lead the solver to explore distinct paths from the same given warm starting point. On instance~\#10, \textit{SPB}-PI achieves 1.081 while the PG methods only reach 0.426--0.444 IQR, indicating that on specific large instances the GPU full warm-start alone can be more effective than partial guidance. 

\subsubsection{Portfolio Extension: Multi-seed Baseline Solver}\label{exp:ablation_portfolio}

Inspired by the multi-seed phase initialization, we want to test whether running $N = 50$ independent solver threads can already improve the performance of the original solvers. To isolate the contribution of GPU guidance, we compare the default single-thread PI vs.\ the default single-thread baseline, and multi-seed PI vs.\ the multi-seed baseline: 50 independent solver threads each with a distinct random seed (as in multi-seed PI).

\begin{table}[ht]
\centering
\vspace{-0.4cm}
\caption{Comparison of GPU-guided phase initialization against the multi-seed portfolio baseline on the 20 best-performing instances.
$\Delta$ reports absolute improvement over the respective single-thread baseline. Values are reported as \textit{TT}/\textit{SPB}.}
\label{tab:portfolio}
\resizebox{\linewidth}{!}{%
\begin{tabular}{l|cc|cc}
\hline
\multirow{2}{*}{\textbf{Configuration}} &
\multicolumn{2}{c|}{\textbf{60\,s}} &
\multicolumn{2}{c}{\textbf{300\,s}} \\
& \textbf{Score} & $\Delta$ & \textbf{Score} & $\Delta$ \\
\hline
\textit{TT}/\textit{SPB} (single-thread)              & 0.527/0.587 & ---/---           & 0.797/0.751 & ---/---           \\
\textit{TT}/\textit{SPB}-PI (single-thread)           & 0.639/0.822 & +0.112/+0.235     & 0.917/0.984 & +0.120/+0.233     \\
\textit{TT}/\textit{SPB} (multi-seed, $N$=50)         & 0.633/0.595 & +0.106/+0.008     & 1.001/0.777 & +0.204/+0.026     \\
\textbf{\textit{TT}/\textit{SPB}-PI (multi-seed, $N$=50)} & \textbf{0.672/0.823} & \textbf{+0.145/+0.236} & \textbf{1.032/1.010} & \textbf{+0.235/+0.259} \\
\hline
\end{tabular}}
\vspace{-0.1cm}
\end{table}

Table~\ref{tab:portfolio} shows that the no-GPU multi-seed portfolio already improves over the single-thread baseline especially for TT ($+$0.204 at 300\,s). Meanwhile, the default single-thread PI delivers substantial GPU-only gains: $+$0.120 for TT-PI and $+$0.233 for SPB-PI at 300\,s. The multi-seed PI then further improves upon single-thread PI by leveraging both GPU guidance and parallel seed diversity. The improvement is especially pronounced for \textit{SPB}: the no-GPU multi-seed baseline reaches the score of 0.777, while the single-thread \textit{SPB}-PI achieves 0.984, and the multi-seed \textit{SPB}-PI reaches 1.010---reinforcing that especially for \textit{SPB}, our GPU warm-start pushes the CPU solver to a high-quality solving path maximizing the benefits of multi-seed.

\subsection{GPU Runtime Overhead}\label{exp:runtime}

As described earlier, our method acts as a complement to the existing SOTA MaxSAT solvers; we still run all experiments and report results at the original solver runtime limit after the GPU overhead. Here we evaluate the GPU overhead across all 229 instances. The GPU pre-solving overhead consists of three stages that run before any CPU solver is launched. Table~\ref{tab:runtime} reports the runtime overhead for each stage, measured over all 229 instances.

\begin{table}[ht]
\centering
\caption{GPU pre-solving overhead per instance over 229 MSE 2024 instances. Timings are solver-independent; all CPU solvers share the same GPU output.}
\label{tab:runtime}
\resizebox{\linewidth}{!}{%
\begin{tabular}{l|l|ccc}
\hline
\textbf{Stage} & \textbf{Description} & \textbf{Avg (s)} & \textbf{Median (s)} & \textbf{Max (s)} \\
\hline
\textbf{(1) Preprocessing}    & 3MaxSAT conversion, tensor gen., GPU load & 34.54 & 6.06 & 553.6 \\
\textbf{(2) Optimization}     & Gradient-based optimization\ (10 epochs)     & 12.76 &  1.70 & 478.5 \\
\textbf{(3) Guidance gen.}    & Sample $N$=50 assignments, save and transfer      &  2.45 &  1.75 &  29.3 \\
\hline
\textbf{Total}                &                                       & \textbf{49.74} & \textbf{9.17} & \textbf{1045.8} \\
\hline
\end{tabular}}
\vspace{-0.1cm}
\end{table}

The GPU pre-solving stage runs as a one-time setup before any CPU solver thread starts. Stage~1 preprocessing---comprising the 3MaxSAT conversion, .pt tensor generation, and GPU tensor load---accounts for the largest portion of the overhead at 34.54\,s on average (69\% of the total), followed by the differentiable optimization in Stage~2 at 12.76\,s (26\%) and guidance generation in Stage~3 at 2.45\,s (5\%). The preprocessing cost (3MaxSAT conversion and tensor generation) is a one-time transformation that can be cached for repeated runs on the same instance. The median total overhead is 9.17\,s, indicating that the GPU total overhead is modest for most instances. The maximum overhead of 1045.8\,s corresponds to instance \texttt{MinimumWeightDominatingSetProblem\_wt-channel-500x100x100-b050} with 4.8M variables; despite this overhead, \textit{TT}-PI achieves a score of 1.004 on this instance at the 300\,s runtime limit---surpassing the best-known MSE 2024 solution---yielding an average gain of $+$0.005 across all four guided configurations. Despite this upfront cost, our GPU guidance consistently pushes the CPU solvers to higher-quality solutions that the backbone baseline solvers alone cannot reach.

%% file: 5-Conclusion.tex
\section{Conclusion}
We presented a GPU-guided framework for Weighted Partial MaxSAT whose key technical contribution is a hierarchical differentiable loss that encodes the hard/soft clause asymmetry directly into the GPU objective, ensuring hard feasibility is always prioritized over soft clause optimization. Two guidance modes transfer the resulting variable assignments to CPU solvers: partial guidance injects high-confidence decisions as soft clauses into $N$ parallel solver instances, while phase initialization seeds the CDCL solver's branching polarity as a warm start without modifying the problem.

Evaluated on all 229 instances of the MSE 2024 anytime weighted benchmark, our framework consistently improves solution quality across both \textit{tt-open-wbo-inc-Glucose} and \textit{SPB-MaxSAT-c-FPS}, achieving up to 4.9\% and 4.8\% average score gains at 300\,s respectively, with meaningful improvements also at 60\,s, with the potential to obtain better solutions than the best recorded solutions, setting a new evaluation standard. Despite modest GPU overhead, our future work is targeting reducing the total runtime, including the preprocessing time. 
The differentiable component provides a training-free, dataset-free approach to instance-specific optimization. Its categorical decision parameters are optimized from the clauses and weights of the current problem, and the resulting assignments guide conventional solvers without requiring a pretrained predictive model.